\documentclass[conference]{IEEEtran}
\usepackage{graphicx}
\usepackage{cite}
\usepackage{amsmath}
\usepackage{pgfplots}
\usepackage{tikz}
\usepackage{subcaption}
\usepackage{caption}
\usepackage{float}

\title{Document Layout Analysis on BaDLAD Dataset: A Comprehensive MViTv2 Based Approach}
\author{
  \IEEEauthorblockN{Ashrafur Rahman}
  \IEEEauthorblockA{\textit{Department of Computer Science} \\
                    \textit{Bangladesh University of Engineering and Technology}\\
                    Dhaka, Bangladesh\\
                    Email: ashrafurkhan37@gmail.com}
  \and
  \IEEEauthorblockN{Asif Azad}
  \IEEEauthorblockA{\textit{Department of Computer Science} \\
                    \textit{Bangladesh University of Engineering and Technology}\\
                    Dhaka, Bangladesh\\
                    Email: asifazad0178@gmail.com}
}

\begin{document}

\maketitle

\begin{abstract}

In the rapidly evolving digital era, the analysis of document layouts plays a pivotal role in automated information extraction and interpretation. In our work, we have trained MViTv2 transformer model architecture with cascaded mask R-CNN on BaDLAD dataset to extract text box, paragraphs, images and tables from a document. After training on 20365 document images for 36 epochs in a 3 phase cycle, we achieved a training loss of 0.2125 and a mask loss of 0.19. Our work extends beyond training, delving into the exploration of potential enhancement avenues. We investigate the impact of rotation and flip augmentation, the effectiveness of slicing input images pre-inference, the implications of varying the resolution of the transformer backbone, and the potential of employing a dual-pass inference to uncover missed text-boxes. Through these explorations, we observe a spectrum of outcomes, where some modifications result in tangible performance improvements, while others offer unique insights for future endeavors.
\end{abstract}


\section{Introduction}
In today's digital age, the exponential growth of textual information has underscored the critical importance of document processing and understanding. Document Layout Analysis (DLA), a fundamental task in the field of computer vision, seeks to unravel the complex structure of documents, enabling automated information extraction and interpretation. At its core, Document Layout Analysis involves decomposing a document into its constituent elements, such as text blocks, paragraphs, images and tables.

However, despite the remarkable strides made in document analysis across various languages, the Bengali language has remained relatively unexplored in this realm. The scarcity of labeled data and dedicated resources has hindered significant advancements in Document Layout Analysis specifically tailored for Bengali documents. In this respect, BaDLAD (Bengali Document Layout Analysis Dataset) is a pioneering effort that has the potential to bridge this very gap and drive the evolution of Bengali document analysis. \cite{shihab2023badlad} As we navigate this landscape, we cast our gaze upon various model architectures that have shaped the field of Document Layout Analysis.

Traditionally, the realm of computer vision has seen the application of Convolutional Neural Networks (CNNs) in document layout analysis. These networks excel at feature extraction from image data, enabling the identification of key structural elements within documents. This approach has yielded valuable insights and paved the way for subsequent advancements.

Region-based Convolutional Neural Networks (R-CNNs) extended the capabilities of CNNs by introducing the concept of region proposals. This enabled more precise localization of elements, a crucial aspect in Document Layout Analysis. R-CNNs offered improved accuracy by focusing on the regions of interest, thereby enhancing the understanding of document structure.

While these approaches demonstrated commendable performance, recent years have witnessed a paradigm shift in the field, with the ascent of the transformer architecture. Originally designed for natural language processing tasks, transformers have proven their versatility by redefining various domains, including Image Segmentation and Document Layout Analysis. The transformer architecture's inherent ability to capture contextual information and long-range dependencies has revolutionized image segmentation. This architectural evolution has spurred the development of models such as the MViT transformer, which has emerged as a state-of-the-art solution in the realm of instance segmentation as well as a potential candidate for Document Layout Analysis tasks.

In the context of BaDLAD, a multi-domain Bengali Document Layout Analysis Dataset, the MViT architecture holds exceptional promise. BaDLAD's diverse collection of documents, spanning multiple domains and layouts, aligns seamlessly with MViT's capacity to capture intricate structural nuances. This marriage of data and architecture positions the MViT transformer as a natural fit for the comprehensive analysis and interpretation of Bengali documents.

\section{Methodology}

\subsection{Model Selection}
Our pursuit of a robust and adept architecture for DLA led us to select the MViTv2-B variant, which has exhibited outstanding performance on the highly regarded COCO dataset. This particular variant, pretrained on IN1k and coupled with the Cascade Mask R-CNN framework, has yielded compelling results. Notably, it achieves a remarkable mask Average Precision (AP) score of 47.4 when evaluated on COCO, following a rigorous training regimen spanning 36 epochs.

The MViTv2-B variant outperforms the Swin-B model by a significant margin, boasting a marked increase of +2.5 and +2.3 in $AP^{box}$ and $AP^{mask}$, respectively. \cite{li2021improved} This remarkable performance enhancement is achieved alongside lower computational demands and a more compact model size, signifying an optimal balance between accuracy and resource efficiency. Hence,we have chosen the MViTv2-B model architecture as the cornerstone of our Document Layout Analysis framework.

\subsection{Preprocessing}
Our preprocessing workflow encompassed several crucial steps to ensure the quality and consistency of input data for our Document Layout Analysis framework. Initially, we implemented color normalization using configuration parameters by aligning pixel mean and standard deviation. To establish uniformity in image dimensions, we conducted image resizing and padding to a 1024$\times$1024 resolution. However, we refrained from binary colorization to preserve the diverse color palettes essential for accurate image element detection.

\subsection{Augmentation}
To increase robustness of training we apply several augmentations on the image inputs. We use random brightness, contrast, saturation and rotation on the train set. These variations simulated real-world scenarios and variations in the dataset. Collectively, these preprocessing steps optimized input data quality, contributing to the reliability and versatility of our Document Layout Analysis across a broad spectrum of document layouts and content.

\subsection{Training}
Our training methodology was designed meticulously, employing a multi-cycle approach to fully harness the potential of the MViTv2-B model for Document Layout Analysis (DLA). Leveraging the Detectron2 framework, we meticulously fine-tuned critical hyperparameters and processes to attain peak performance. \cite{wu2019detectron2}

Our training was strategically organized into three consecutive cycles, each spanning 12 epochs for a total of 36 epochs. The model's parameters were efficiently transferred between training cycles, with the final parameters of each 12-epoch run serving as the starting point for the subsequent run. This approach enabled us to explore different training and model parameters and compare them between training phases.

Optimization was achieved through the AdamW optimizer, with a initial learning rate of $8*10^{-5}$. The learning rate multiplier scheduler was meticulously tailored to guide the model's convergence over different iteration ranges. During the initial 50 iterations a warmup phase was employed with a warmup factor of $10^{-3}$, starting with a learning rate of $8*10^{-8}$ and gradually upto $8*10^{-5}$ after the warmup phase. We started with a slow learning rate with the intention to avoid overshooting and promote model stability. Moreover, the learning rate was further decreased by $0.1$ times and $0.01$ times at about $88\%$ and $97\%$ of the total iterations respectively to make small adjustments in the final iterations and aid the model's convergence.

We structured data loading and processing using a batch size of 16 for training. During training, crucial metrics were logged every 20 iterations, and model checkpoints were saved every 2000 iterations to facilitate potential model recovery.

\section{Results}

After a total training of 36 epochs in 3 phases, our model scored a DICE score of 0.90095 on the public test set and a total loss of 0.2125 and a mask loss of 0.19. The loss metrics gradually decreased throughout the training. \par
\begin{figure}[H]
\begin{tikzpicture}[>=latex]
    \centering
    \begin{axis}[
      axis x line=center,
      axis y line=center,
      width={\linewidth},
      xtick={1,5,10,15,20,25,30,35},
      ytick={0.1, 0.2, 0.3, 0.4, 0.5},
      xlabel={epoch},
      ylabel={loss\_mask},
      xlabel style={right},
      ylabel style={above},
      xmin= 0,
      xmax=38,
      ymin=0.1,
      ymax=0.55]
    
      \addplot+ [color=black,mark=none] table {
        1 0.48
        2 0.41
        3 0.405
        4 0.35
        5 0.33
        6 0.3
        7 0.32
        8 0.28
        9 0.24
        10 0.25
        11 0.235
        12 0.22
        13 0.215
        14 0.21
        15 0.23
        16 0.21
        17 0.23
        18 0.22
        19 0.21
        20 0.205
        21 0.22
        22 0.215
        23 0.22
        24 0.21
        25 0.205
        26 0.198
        27 0.21
        28 0.205
        29 0.2077
        30 0.21
        31 0.202
        32 0.21
        33 0.198
        34 0.204
        35 0.18
        36 0.19
    };
    \end{axis}
\end{tikzpicture}
\caption{Loss mask vs Epoch}
\label{fig:mask_loss}
\end{figure}
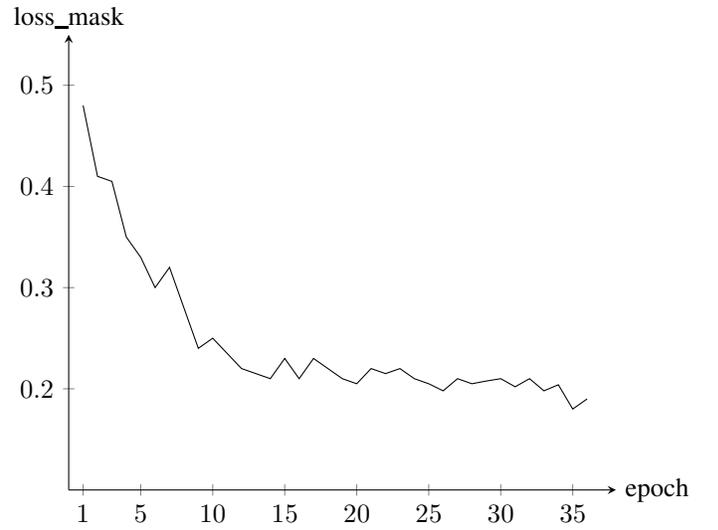

 
From figure \ref{fig:mask_loss} we can see that the mask loss reached about 0.2 after the first 12 epochs. However, the training loss slowly continued to improve over the training period.

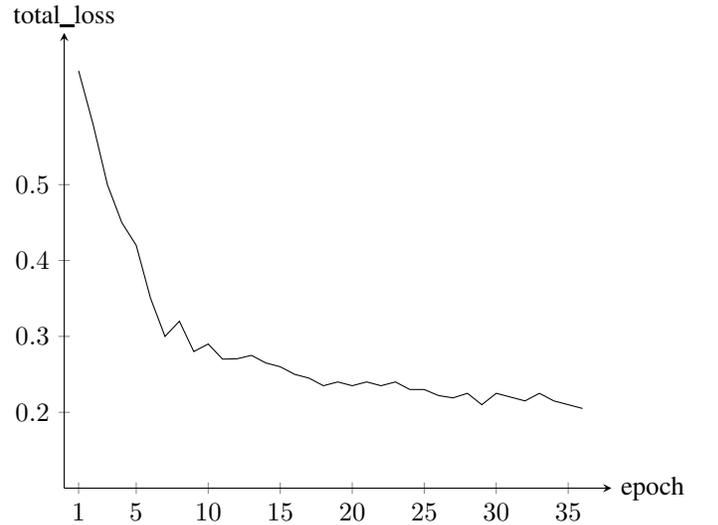
\begin{figure}[H]
\begin{tikzpicture}[>=latex]
    \centering
    \begin{axis}[
      axis x line=center,
      axis y line=center,
      width={\linewidth},
      xtick={1,5,10,15,20,25,30,35},
      ytick={0.1, 0.2, 0.3, 0.4, 0.5},
      xlabel={epoch},
      ylabel={total\_loss},
      xlabel style={right},
      ylabel style={above},
      xmin= 0,
      xmax=38,
      ymin=0.1,
      ymax=0.7]
    
      \addplot+ [color=black,mark=none] table {
        1 0.65
        2 0.58
        3 0.5
        4 0.45
        5 0.42
        6 0.35
        7 0.3
        8 0.32
        9 0.28
        10 0.29
        11 0.27
        12 0.2705
        13 0.275
        14 0.265
        15 0.26
        16 0.25
        17 0.245
        18 0.235
        19 0.24
        20 0.235
        21 0.24
        22 0.235
        23 0.24
        24 0.23
        25 0.23
        26 0.222
        27 0.219
        28 0.225
        29 0.21
        30 0.225
        31 0.22
        32 0.215
        33 0.225
        34 0.215
        35 0.21
        36 0.205
    };
    \end{axis}
\end{tikzpicture}
\caption{Total loss vs Epoch}
\label{fig:total_loss}
\end{figure}

Table \ref{table:scores} summarizes the scores obtained after each phase of training.

\begin{figure}[H]
    \centering
    \begin{tabular}{|c|c|c|}
       \hline 
       Epoch  & Training loss & Dice Score \\
       \hline 
       12  & 0.2705 & 0.89904 \\
       24  & 0.2322 & 0.90244 \\
       36  & 0.2078 & 0.90348 \\
       \hline
    \end{tabular}
    \caption{Test Set Scores}
    \label{table:scores}
\end{figure}

\section{Kaggle Competition Results}

In the Kaggle competition "DL Sprint 2.0," which focused on Document Layout Analysis using the BaDLAD Dataset, a total of 94 teams participated. Our team, named "Black Quad," demonstrated exceptional performance and secured the championship.

Our approach, titled "Document Layout Analysis on BaDLAD Dataset: A Comprehensive MViTv2 Based Approach," yielded remarkable results. We achieved the highest Dice score in the competition of 0.90396, showcasing the effectiveness of our method in accurately segmenting document layouts. 

Furthermore, our team achieved a mean Average Precision (mAP) score of 56.381, which stands out as the highest among all participating teams. This accomplishment underscores the robustness and generalization of our approach across various document layouts and scenarios.

\section{Discussion}

\subsection{Effect of Rotation and Flip}
We observed that both the training set and test set contains rotated images. Initially, in order to accommodate for this case, we introduced random discreet rotations from the set $\{ 0^{\circ}, 90^{\circ}, 180^{\circ}, 270^{\circ} \}$. However we observed that this augmentation as well as horizontal or vertical flips lead to poorer results. Since rotated images are rare in the test set, we chose to train our model to primarily focus on upright images. A more comprehensive approach would be to train a separate model to recognize rotations and preprocess them during inference. \par
However, We noticed a much more prevalent pattern of rotations in the dataset. Since much of the documents are scanned documents, they are slightly tilted at a small angle. To incorporate this variation we augmented our images using a random rotation in the range $[-5^{\circ}, 5^{\circ}]$.

\subsection{Sliced Inference}
During our evaluation of the inference of our model both on the train and public test set we observed that many small features, primarily text-boxes and paragraphs weren't recognized with adequate confidence by our model. This observation initially led us to believe that slicing the input images into overlapping windows and running inference on each of the slices would improve our results furthermore. \par 
However, despite a two to four fold increase in inference time, we didn't observe any noticeable improvement of the model's capability of recognizing smaller instances. Slicing also introduced the problem of recognising larger features such as tables or images in multiple segments.

\subsection{Resolution of Transformer Backbone}
The multiscale vision transformer backbone of the MViTv2 model can work in different resolutions. \cite{li2021improved} The pretrained models are trained on image resolution of $224\times224$. We first trained 12 epochs on the pretrained MViTV2-B model with $224\times224$ resolution. We then trained the model at $384\times384$ resolution, initializing weights from the fine tuned $224\times224$ model. We repeated this process for resolution $512\times512$, using the weights from the previous step. We observed that with same similar training the $224\times224$ resolution model performs better.

\subsection{Two pass Inference to Detect Missed Text-boxes}
In an attempt to recognize the smaller text-boxes and paragraphs missed during inference, we designed a two pass inference approach. We noted that overcrowding of instances played a big factor in the model's failure to detect some features. Therefore, to identify those features, we ran inference twice. The recognised text-boxes and paragraphs, from the first inference, which didn't overlap with any images were erased (i.e. replaced with the background colour). The resulting image was then fed to the model for a second inference. The text-boxes recognized in this pass were added to the model's prediction result. However, this approach was found to perform worse than the single pass inference on the public test set.

\subsection{Effectiveness of Transformer Model}
While most other solutions explored traditional CNN based models including R50-FPN and yolov8, we explored several transformer based architectures including maskDINO and MViTv2. We found both models show improved performance after similar training iterations. The MViTv2-B model gave us a balance of performance and resource efficiency as noted earlier, increasing accuracy compared to traditional approaches while not severely increasing inference time or memory consumption.

\section{Conclusion and Future Work}
We present an effective scheme to fine tune the transfermer based MViTv2 model for Bengali Document Layout Analysis. Previously traditional CNN based models were applied on the dataset.\cite{shihab2023badlad} Our work demonstrates the viability of the transformer based Vision Transformer models such as MViTv2 for Bengali Document Layout Analysis. \par

As noted before, handling rotation explicitly by training a separate lightweight model to recognize rotations will increase the model's capacity to handle arbitrarily rotated images. Another limitation of the model is it's failure to detect small objects, mainly text-boxes. Copy-paste augmentation techniques have proven to significantly improve a model's capacity in these regards\cite{ghiasi2021simple, kisantal2019augmentation}. We couldn't train our model with copy-paste augmentation due to shortage of time but we believe incorporating this augmentation will further strengthen our model and increase it's robustness.

% References section

\bibliographystyle{IEEEtran}
% Generated by IEEEtran.bst, version: 1.14 (2015/08/26)

% \bibliography{references}

\end{document}